\documentclass{article} 
\usepackage{iclr2025_conference,times}

\usepackage{hyperref}
\usepackage{url}

\usepackage{times}
\usepackage{latexsym}
\usepackage[utf8]{inputenc}
\usepackage[T1]{fontenc}

\usepackage{microtype}

\usepackage{inconsolata}

\usepackage{graphicx}

\usepackage{float}
\usepackage{booktabs}
\usepackage{multirow}

\usepackage{svg}
\usepackage{amsmath} 
\DeclareMathOperator*{\argmax}{arg\,max}

\usepackage{caption}
\usepackage{subcaption}
\usepackage{tikz}
\usepackage{pgfplots}
\usepgfplotslibrary{fillbetween}
\pgfplotsset{compat=1.16}
\usepackage{multicol}

\newenvironment{customlegend}[1][]{%
    \begingroup
    \csname pgfplots@init@cleared@structures\endcsname
    \pgfplotsset{#1}%
}{%
    \csname pgfplots@createlegend\endcsname
    \endgroup
}%

\def\addlegendimage{\csname pgfplots@addlegendimage\endcsname}
\definecolor{acolor}{rgb}{1.0, 0.5490196078431373, 0.0}
\definecolor{bcolor}{rgb}{0.5019607843137255, 0.0, 0.5019607843137255}
\definecolor{ccolor}{rgb}{0.0, 0.5019607843137255, 0.0}

\definecolor{dcolor}{rgb}{0, 0.302, 0.596}
\definecolor{ecolor}{rgb}{0.647, 0, 0.267}
\definecolor{fcolor}{rgb}{0, 0, 0}
\usepackage{microtype}

\title{Quasi-random Multi-Sample Inference for Large Language Models}

\author{Avinash Amballa\thanks{ 
indicates equal contribution} , Aditya Parashar\footnotemark[1] , Aditya Vikram Singh\footnotemark[1] , Jinlin Lai , Benjamin Rozonoyer \\
\hspace{5mm} College of Information \& Computer Sciences, University of Massachusetts Amherst\\
\hspace{10mm} \texttt{\{aamballa, aparashar, 
avsingh, jinlinlai, brozonoyer\}@umass.edu} \\
}

\iclrfinalcopy 
\begin{document}

\maketitle

\begin{abstract}

Large language models (LLMs) are often used with decoding strategies that require sampling multiple outputs. \citet{vilnis2023arithmetic} show that an LLM implicitly defines an arithmetic code book, facilitating efficient and embarrassingly parallelizable \textbf{arithmetic sampling} to produce multiple samples using quasi-random codes. Traditional text generation methods, such as beam search and sampling-based techniques, have notable limitations: they lack parallelizability or diversity of sampled sequences. This study explores the potential of arithmetic sampling, contrasting it with ancestral sampling across two decoding tasks that employ multi-sample inference: chain-of-thought reasoning with self-consistency and machine translation with minimum Bayes risk decoding. Our results demonstrate that arithmetic sampling produces more diverse samples, significantly improving reasoning and translation performance as the sample size increases. We observe a $\mathbf{3\text{-}5\%}$ point increase in accuracy on the GSM8K dataset and a $\mathbf{0.45\text{-}0.89\%}$ point increment in COMET score for WMT19 tasks using arithmetic sampling without any significant computational overhead.
\end{abstract}

\section{Introduction}

There have been enormous efforts in improving the performance and efficiency of inference with large language models \citep{ippolito2019comparison,su2023contrastive,grubisic2024priority,zhou2024survey,ding2024efficiency} based on system, data, and model level enhancements. 
In this paper, we consider that any decoding routine can be broadly assessed by its sample diversity (\ref{app: A1}) and parallelizability. Search-based techniques like beam search can approximate \textit{maximum a posteriori} (MAP) decoding, mitigating duplicate samples at the expense of not being embarrassingly parallel. Sampling-based methods grounded in ancestral sampling techniques are parallel but don't explicitly guarantee diverse sequences. The recently proposed arithmetic sampling \cite{vilnis2023arithmetic} technique enables parallel inference with diverse samples -- by interpreting the inference as sampling from code points from a unit interval, given code points generating sequences becomes embarrassingly parallel and the sample diversity is guaranteed by construction.

Decoding from pre-trained LLMs requires varying strategies for different downstream tasks. For complex reasoning and question-answering tasks, chain-of-thought (CoT) prompting \cite{wei2022chain} is established for improving inference by instructing the model to generate intermediate reasoning paths. \citet{wang2023selfconsistency} propose self-consistency as an additional improvement over chain-of-thought reasoning with multi-sample inference, attributable to diverse reasoning paths enhancing the confidence of the majority answer. For machine translation, minimum Bayes risk (MBR) decoding \cite{kumar2004minimum} is a classical approach for selecting the optimal translation from candidate translations generated by an LLM, requiring diversity to ensure performance.

Thus, the inherent diversity of sequences generated via arithmetic sampling offers significant potential for enhancing decoding strategies that rely on multi-sample inference. Recognizing the importance of exploring this approach, we apply arithmetic sampling to both reasoning and translation tasks. For CoT reasoning with self-consistency and machine translation with MBR decoding, we observe accuracy improvements on the GSM8K and Commonsense QA datasets, along with substantial COMET score gains as the number of sampled sequences increases. 

\begin{figure*}[h]
\centering
\resizebox{\linewidth}{!}{%
    \begin{tabular}{ccc}
        \captionsetup[subfloat]{font=huge,labelfont=huge}
        \subfloat[Accuracy vs. \# sampled sequences: Gemma-7B]{{\includegraphics[width=\linewidth]{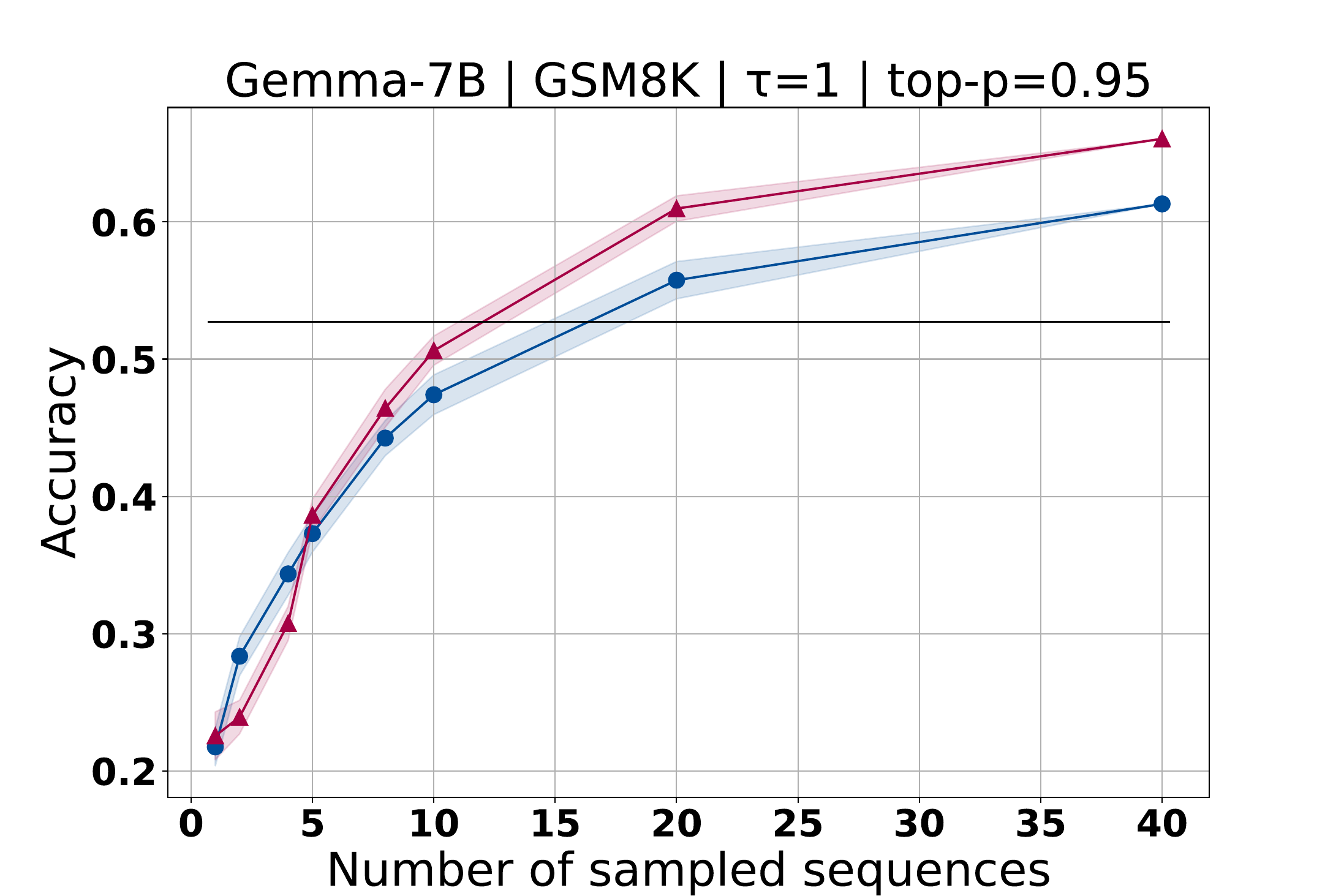}}} &
        \captionsetup[subfloat]{font=huge,labelfont=huge}
        \subfloat[Accuracy vs. \# sampled sequences: Llama-2-7B]{{\includegraphics[width=\linewidth]{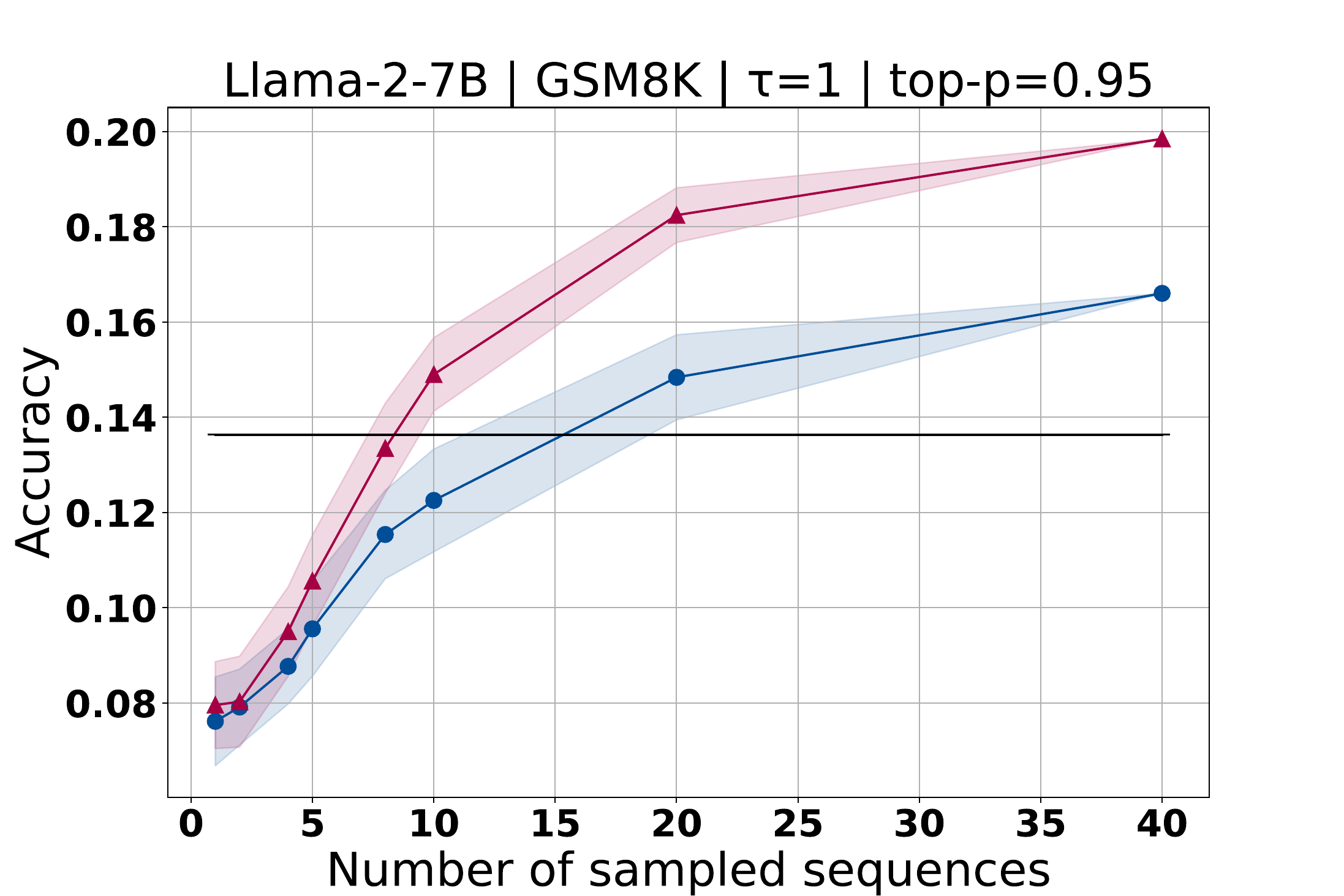}}} &
         \captionsetup[subfloat]{font=huge,labelfont=huge}
        \subfloat[Accuracy vs. n-gram diversity: Llama-2-7B]{{\includegraphics[width=\linewidth]{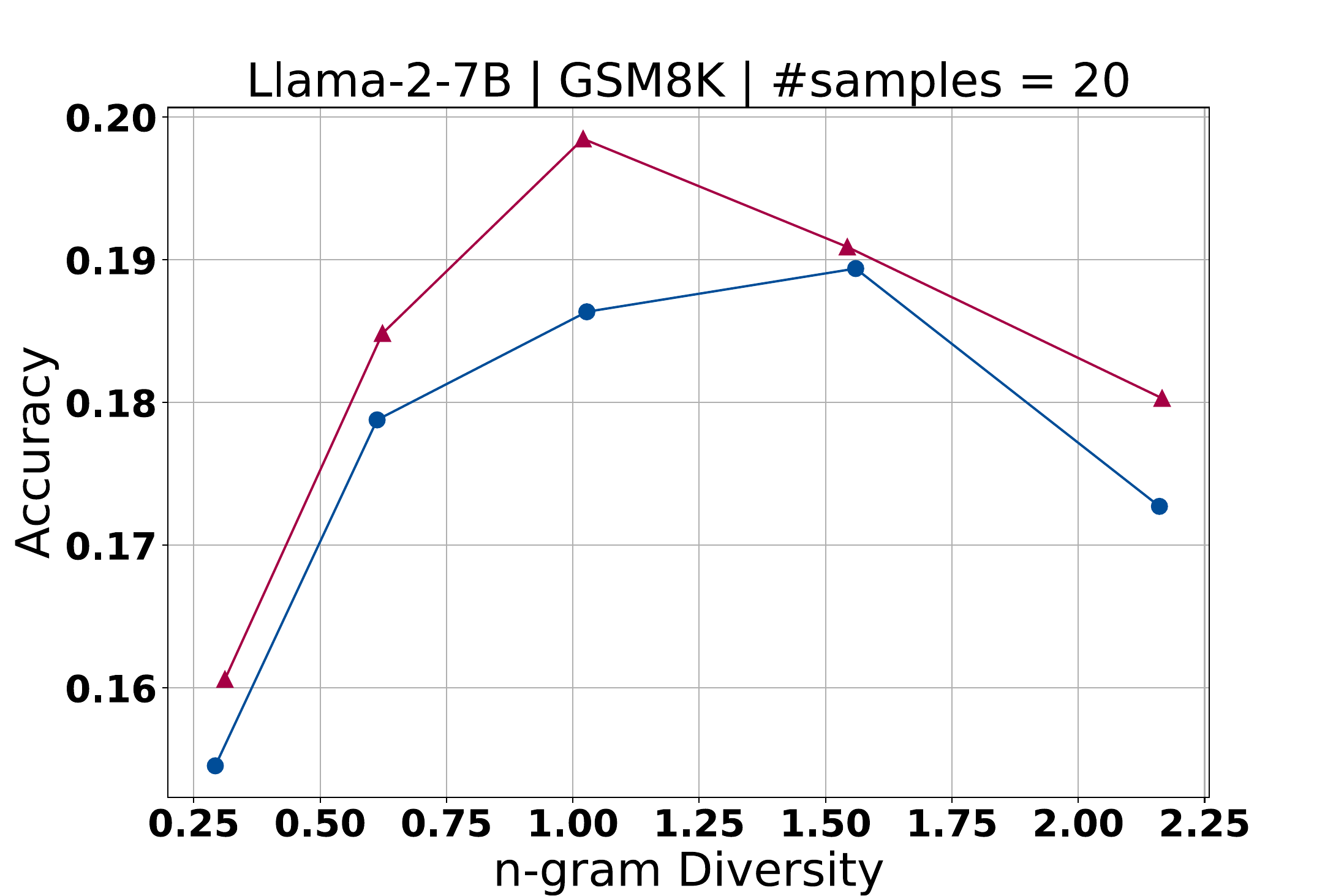}}}
    \end{tabular}
    }
    \caption{8-shot evaluation on GSM8K with Gemma-7B and Llama-2-7B}
    \label{fig:gsm8k_analysis}
    \vspace{-10pt}
\end{figure*}
\begin{figure*}[h]
\centering
\resizebox{\linewidth}{!}{%
    \begin{tabular}{ccc}
        \captionsetup[subfloat]{font=huge,labelfont=huge}
        \subfloat[Accuracy vs. \# sampled sequences: Gemma-7B]{{\includegraphics[width=\linewidth]{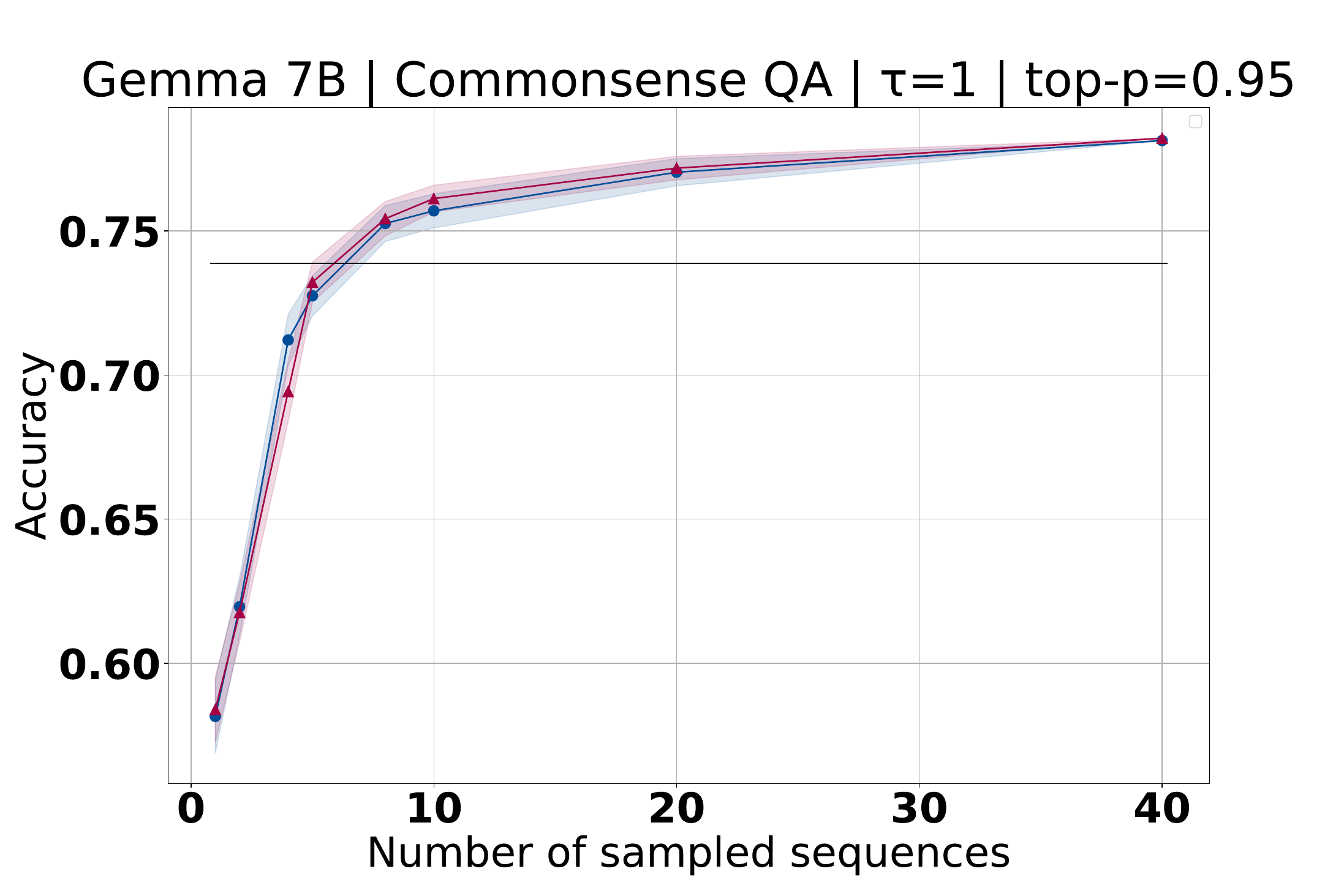}}} &
        \captionsetup[subfloat]{font=huge,labelfont=huge}
        \subfloat[Accuracy vs. \# sampled sequences: Llama-2-7B]{{\includegraphics[width=\linewidth]{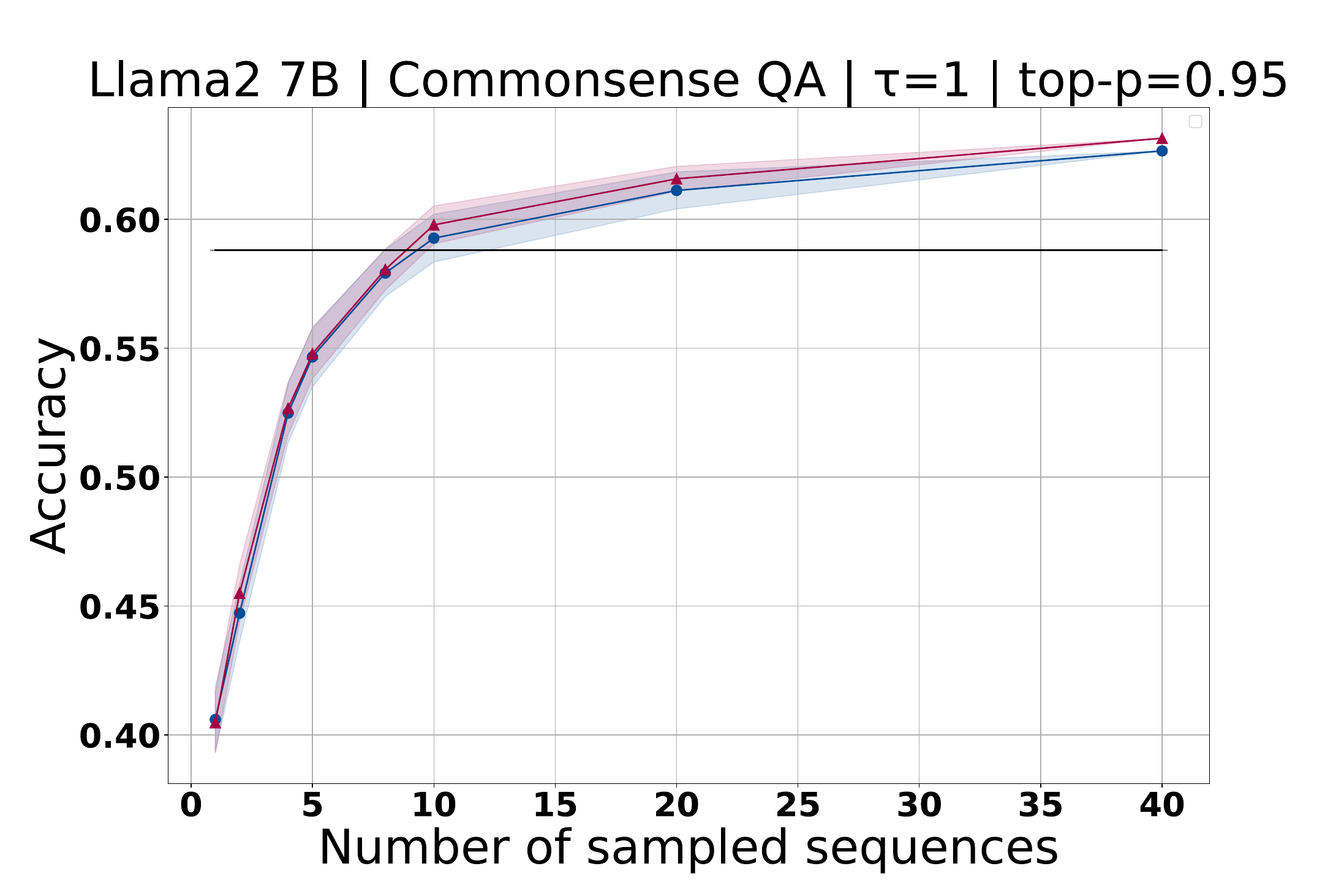}}} &
         \captionsetup[subfloat]{font=huge,labelfont=huge}
        \subfloat[Accuracy vs. n-gram diversity: Llama-2-7B]{{\includegraphics[width=\linewidth]{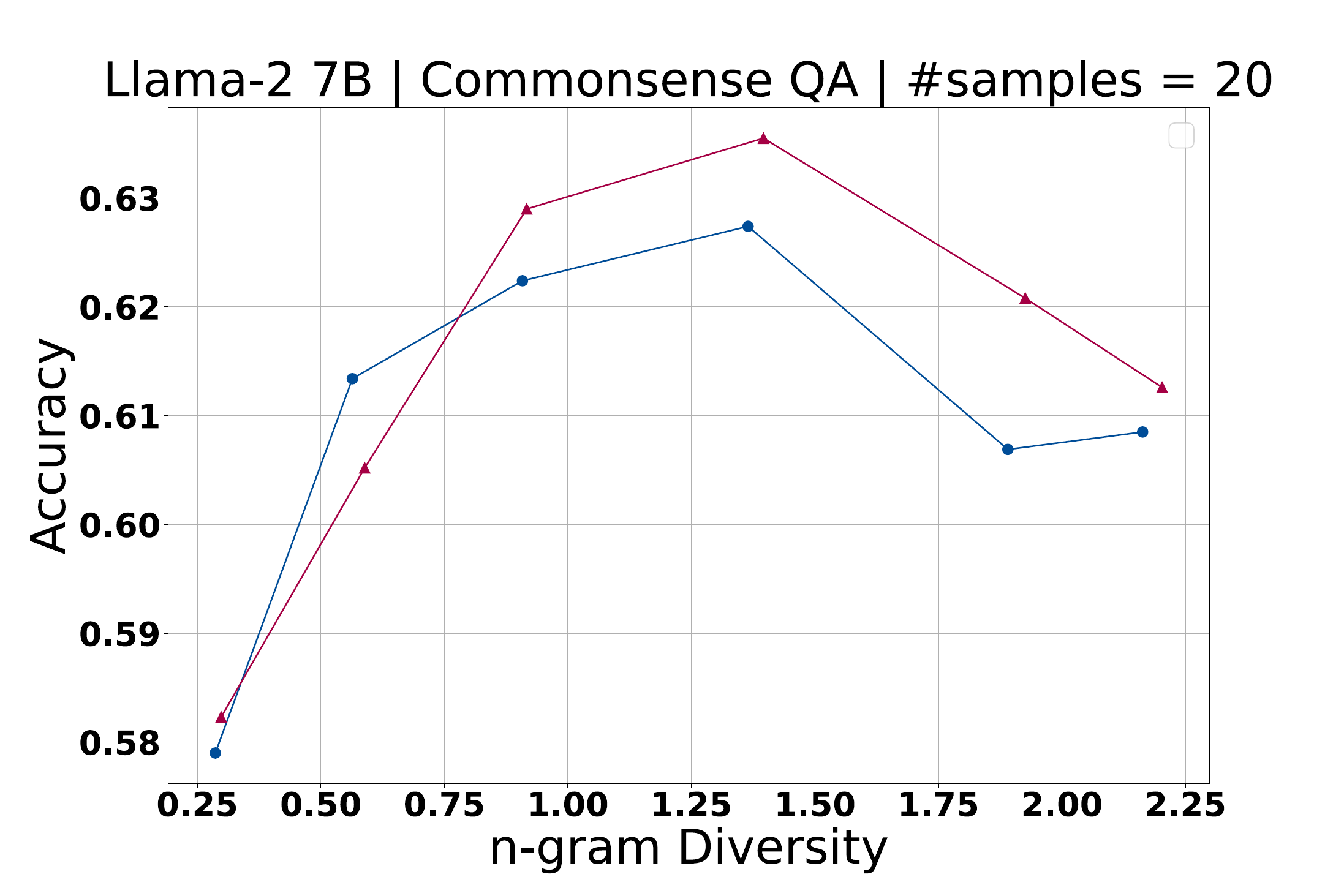}}}
    \end{tabular}
    }
    \begin{tikzpicture}
        \begin{customlegend}[
            legend columns=3,
            legend style={
                align=left,
                column sep=2ex
            },
            legend entries={Ancestral sampling, Arithmetic sampling, Greedy}
        ]
            \addlegendimage{mark=*,solid,color=dcolor}
            \addlegendimage{mark=triangle*,mark size=3pt,solid,color=ecolor}
            \addlegendimage{mark=circle*,solid,color=fcolor}
        \end{customlegend}
    \end{tikzpicture}   
    \caption{6-shot evaluation on Commonsense QA with Gemma-7B and Llama-2-7B}
    \label{fig:csqa_analysis}
    \vspace{-10pt}
\end{figure*}
\section{Background}
\subsection{Arithmetic sampling} Arithmetic sampling \cite{vilnis2023arithmetic} reinterprets the standard ancestral sampling process as lazily constructing an arithmetic codebook in the unit $[0, 1)$ interval where each code (point) is uniformly distributed and corresponds to a sequence from the output distribution.

This process ensures that the generated samples are diverse, as codes far apart in the code book usually correspond to different token prefixes. Moreover, it is embarrassingly parallel across the $N$ samples, since the $i^\text{th}$ sample can be generated independently given its code $c_i$. Arithmetic sampling can also be applied orthogonally to other sampling-based techniques that directly manipulate the next token distribution, such as top-k, top-p (nucleus) \cite{holtzman2020curious},  temperature sampling and epsilon sampling \cite{hewitt-etal-2022-truncation}.

\subsection{Self-consistency} Self-consistency \cite{wang2023selfconsistency} is a method designed to improve the performance of chain-of-thought (CoT) reasoning by generating multiple reasoning paths with answers for a given prompt and then selecting the most consistent answer based on, generally, majority voting. This approach leverages the diversity of the generated candidate reasoning paths to identify the most frequent outcome, thereby enhancing the accuracy of the final answer.

\subsection{MBR decoding} Minimum Bayes risk (MBR) \cite{eikema2020map} is based on the principle of maximizing the expected utility of a given hypothesis. When making predictions, we lack information about the ideal (target) translations and must make decisions under uncertainty. MBR allows the model to probabilistically estimate ideal decisions as it searches for the candidate that maximizes expected utility. We used COMET \cite{rei2020cometneuralframeworkmt} as the utility metric.

We use the sampling-based approximation to MBR decoding as posited in \cite{eikema-aziz-2022-sampling}, using the Monte Carlo estimate and formulating the candidate space from the generated $N$ (pseudo-reference) samples:
\vspace{-10pt}
\[
y^{\text{N-by-N}} =  \underset{h \in \{y^{(1)}, y^{(2)}, \dots y^{(N)} \}}{\operatorname{\argmax}} \frac{1}{N} \sum_{n=1}^{N} u(y^{(n)}, h)
\]
\vspace{-10pt}


\begin{figure*}[h]
\centering
\resizebox{0.8\linewidth}{!}{%
    \begin{tabular}{cccc}
        \captionsetup[subfloat]{font=Huge,labelfont=Huge}
        \subfloat[De-En task: Flan-T5]{{\includegraphics[height=10cm]{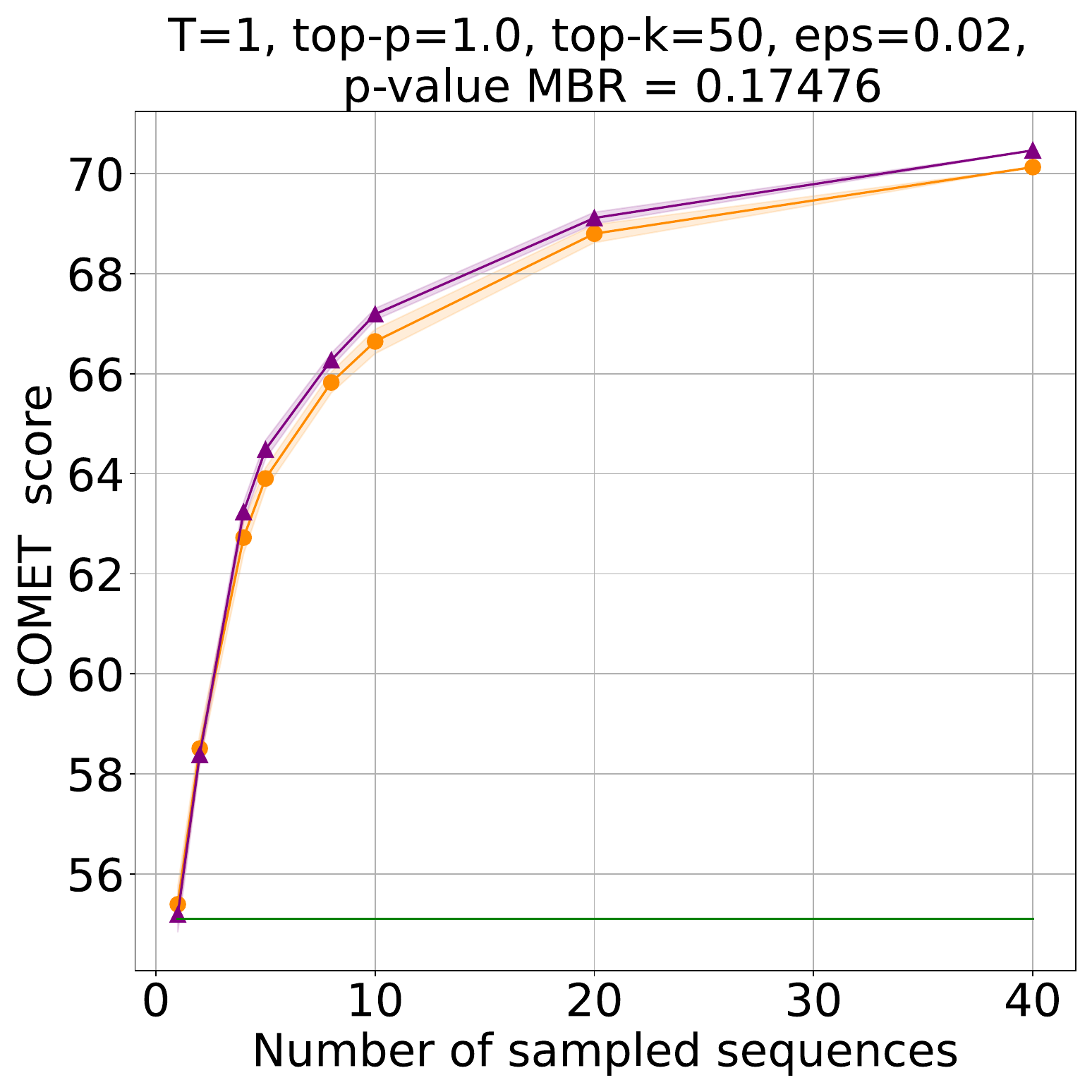} }} &
        \captionsetup[subfloat]{font=Huge,labelfont=Huge}
        \subfloat[De-En task: MT0 ]{{\includegraphics[height=10cm]{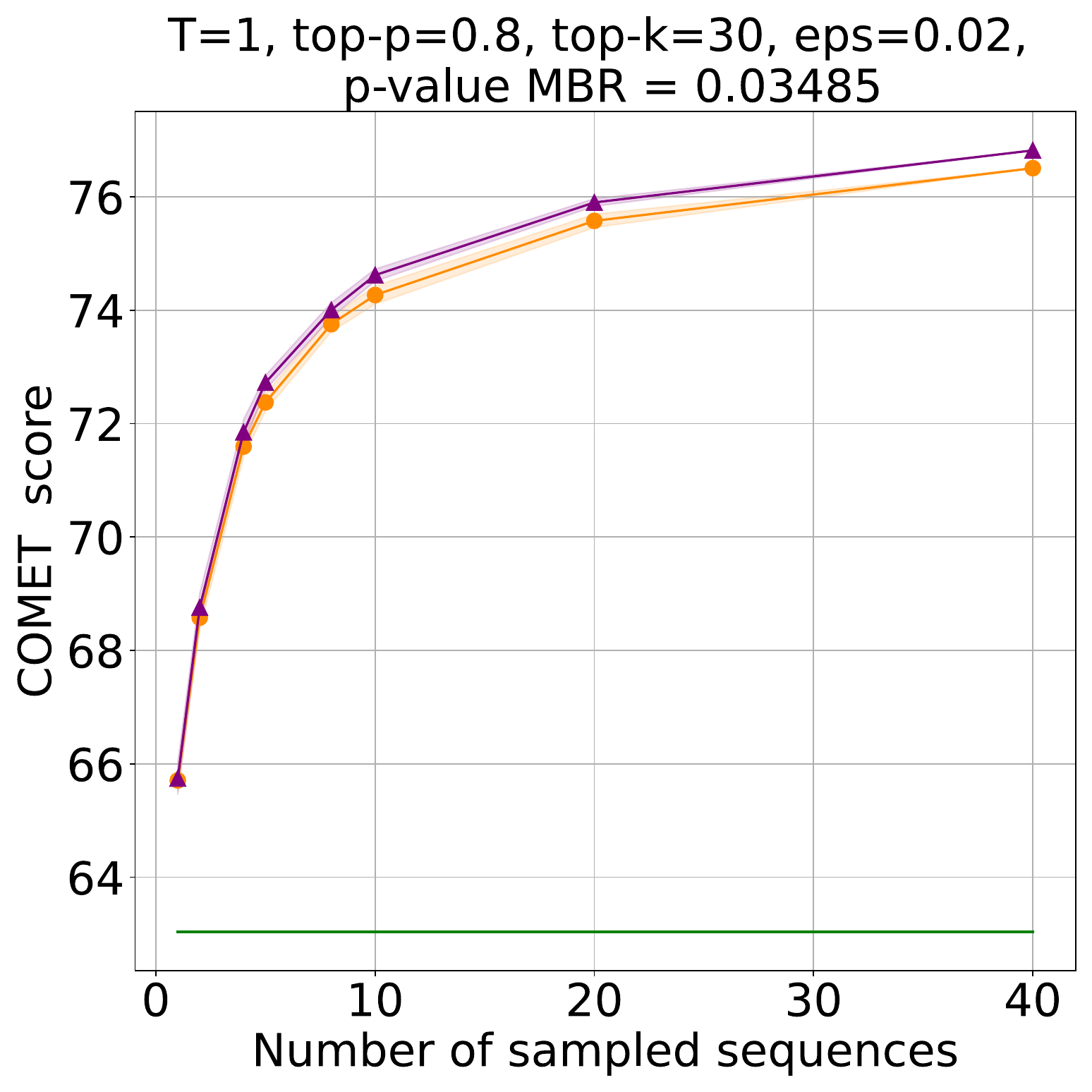} }} &
        \captionsetup[subfloat]{font=Huge,labelfont=Huge}
        \subfloat[Ru-En task: MT0 ]{{\includegraphics[height=10cm]{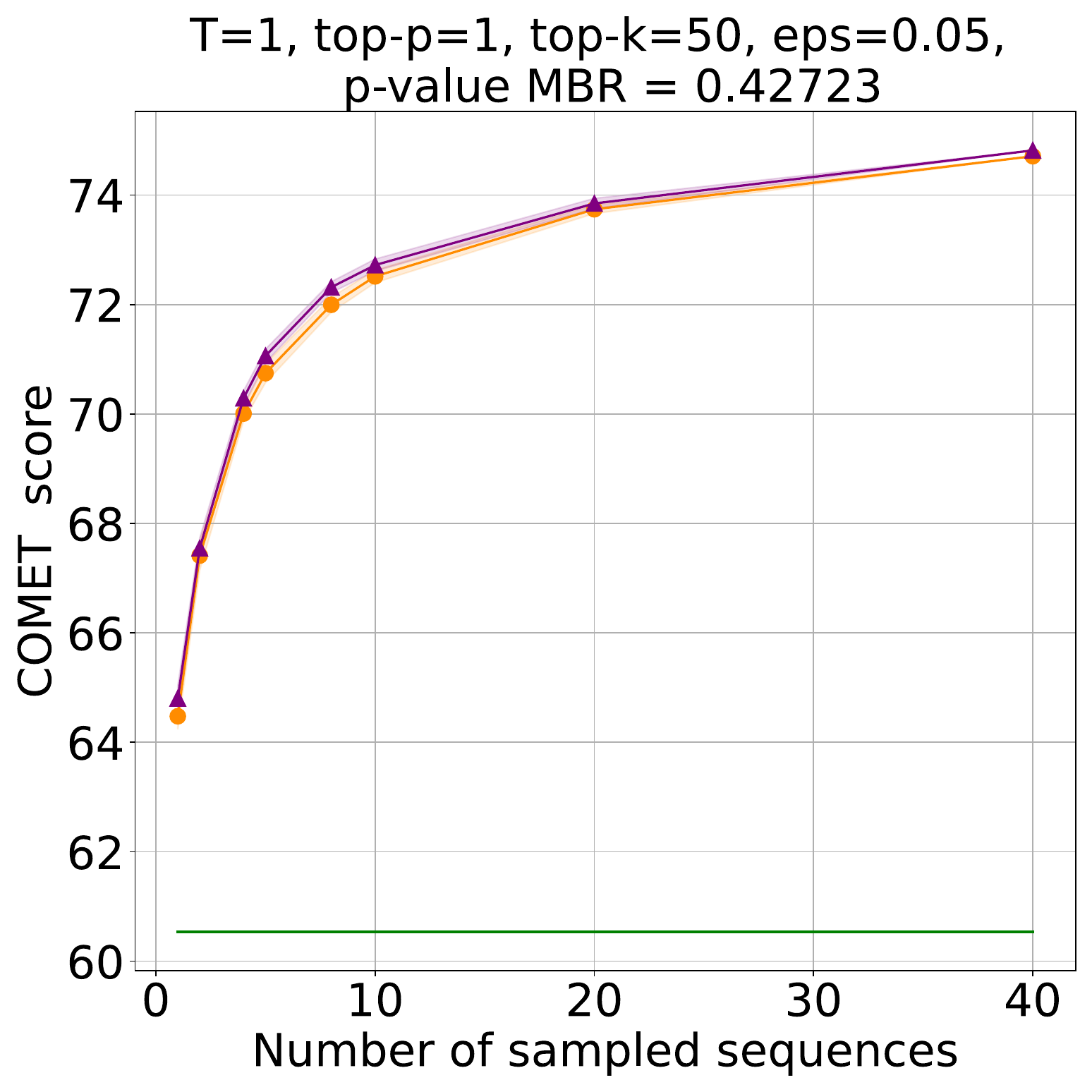} }} &
    \end{tabular}
    }
    \caption{COMET vs. \#sampled sequences on Flan T5, MT0}
    \label{fig:MBR1}
    \vspace{-10pt}
\end{figure*}

\begin{figure*}[h]
\centering
\resizebox{0.8\linewidth}{!}{%
    \begin{tabular}{cccc}
         \captionsetup[subfloat]{font=Huge,labelfont=Huge}
        \subfloat[De-En task: Flan-T5 ]{{\includegraphics[height=10cm]{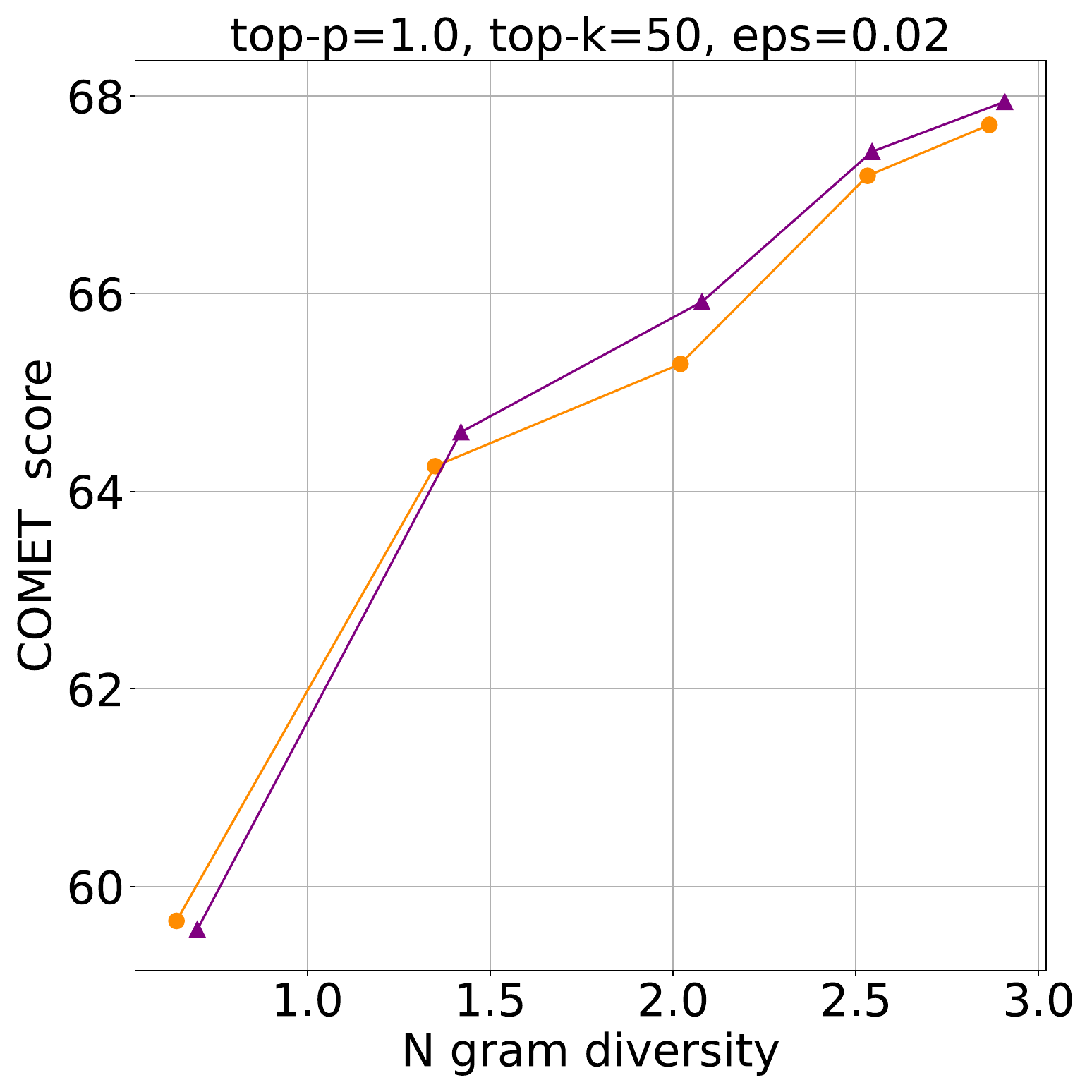} }} & 
         \captionsetup[subfloat]{font=Huge,labelfont=Huge}
        \subfloat[De-En task: MT0 ]{{\includegraphics[height=10cm]{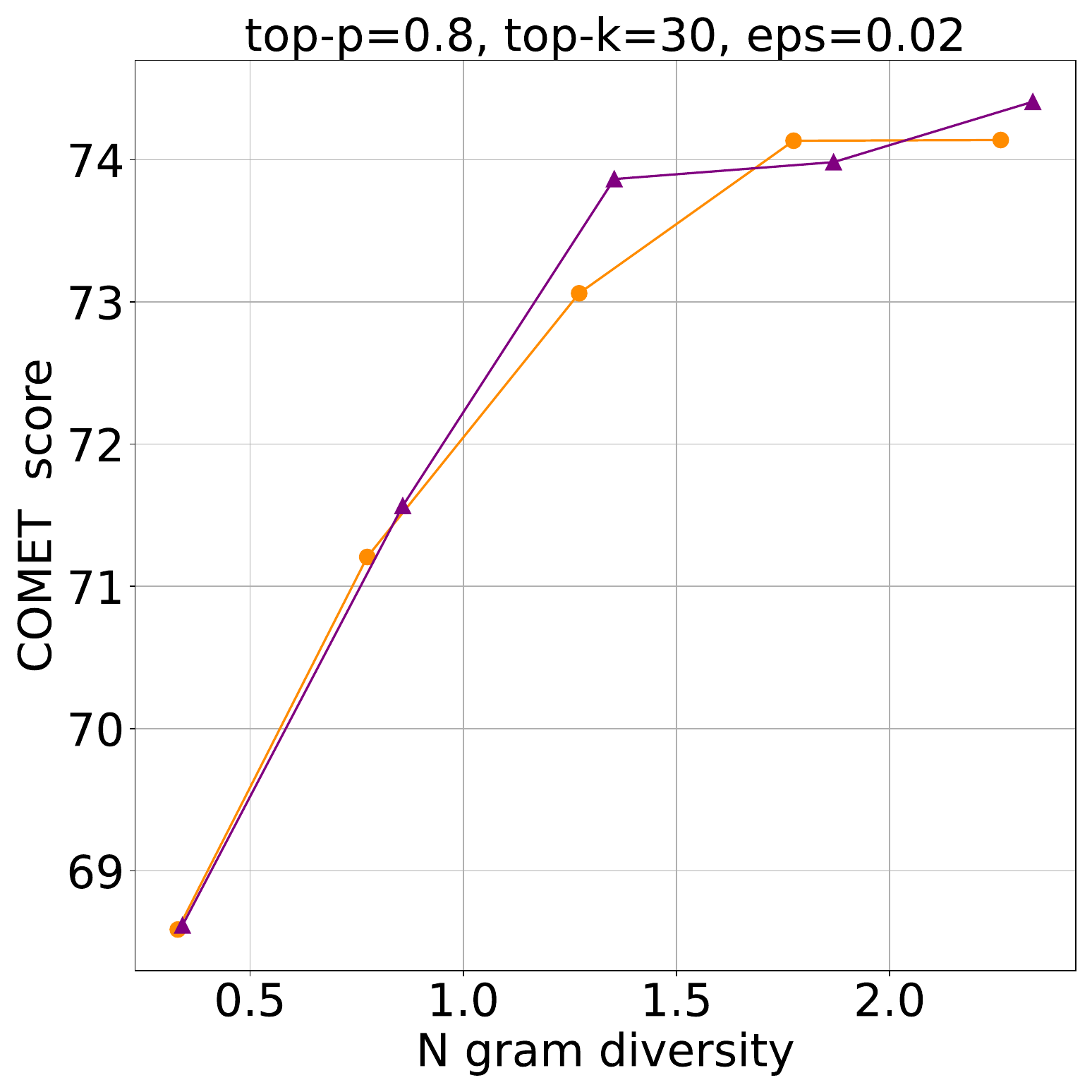} }}  & 
         \captionsetup[subfloat]{font=Huge,labelfont=Huge}
        \subfloat[Ru-En: MT0]{{\includegraphics[height=10cm]{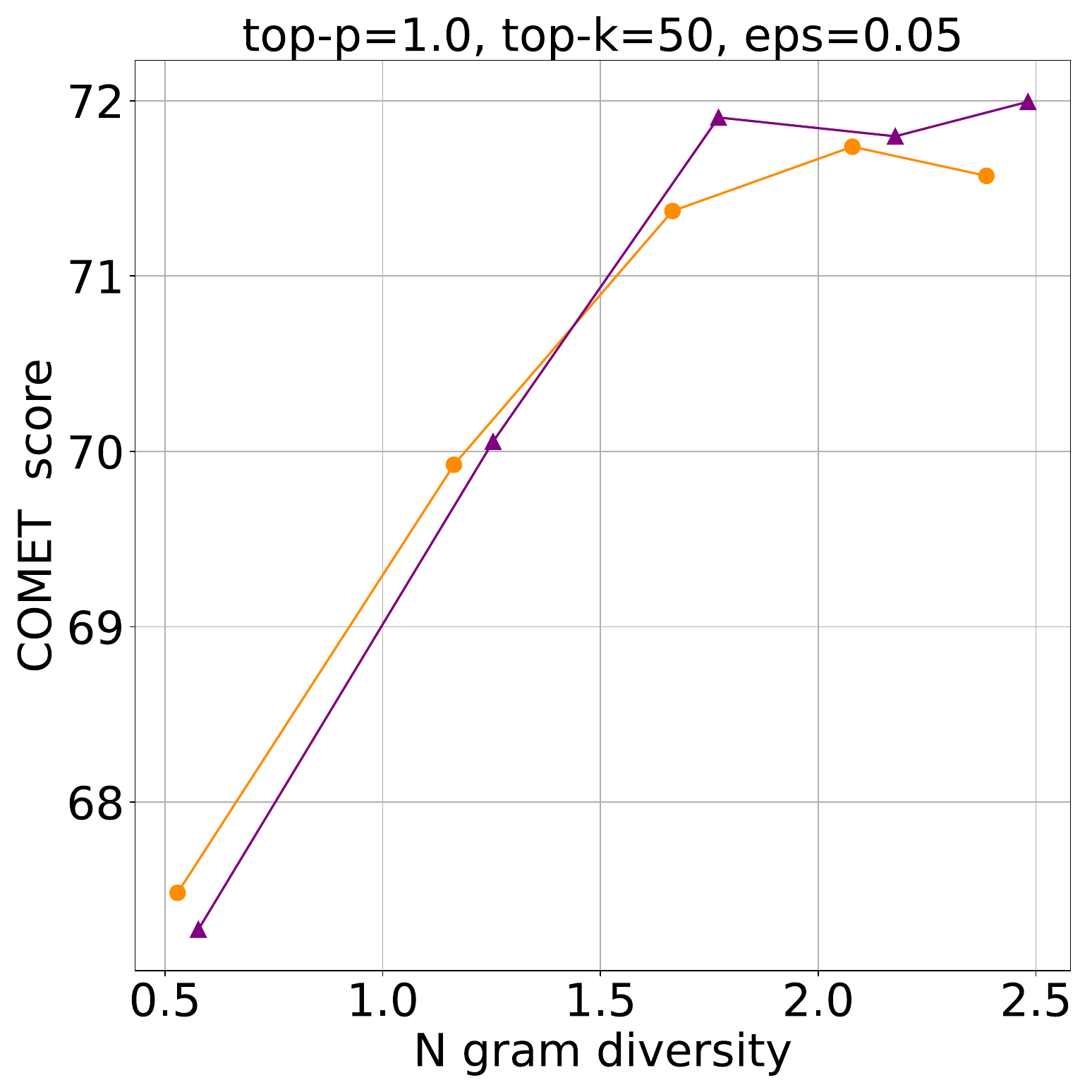} }} 
    \end{tabular}
    }
      \begin{tikzpicture}
        \begin{customlegend}[
            legend columns=3,
            legend style={
                align=left,
                column sep=2ex
            },
            legend entries={ Ancestral sampling (MBR), Arithmetic sampling (MBR), Greedy}
        ]
            \addlegendimage{mark=*,solid,color=acolor}
            \addlegendimage{mark=triangle*,mark size=3pt,solid,color=bcolor}
            \addlegendimage{mark=circle*,solid,color=ccolor}
        \end{customlegend}
    \end{tikzpicture}   
    \caption{COMET vs. n-gram diversity on Flan-T5, MT0 varying temperature $T$}
    \label{fig:MBR2}
    \vspace{-10pt}
\end{figure*}
\section{Experiments}
\subsection{Datasets}
We conduct experiments with only validation splits of datasets aimed at chain-of-thought reasoning and neural machine translation. We use  \textbf{GSM8K} \cite{gsm8k} and \textbf{CommonsenseQA} \cite{commonsenseqa} for chain-of-thought self-consistency; and \textbf{WMT19 De -- En (German-English)} and \textbf{Ru-En (Russian-English)} \cite{barrault-etal-2019-findings} for MBR decoding machine translation experiments. 

\subsection{Models and Baselines}
In our experiments, we employ Gemma-7B \cite{gemmateam2024gemma}, and Llama-2-7B \cite{touvron2023llama} models for CoT self-consistency; and MT0 \cite{muennighoff2023crosslingual} for De-En and Ru-En tasks, Flan-T5 \cite{chung2022scaling} for De-En task with MBR decoding.

In addition to greedy decoding, we adopt ancestral sampling as our baseline method to compare with arithmetic sampling for sampling reasoning paths in self-consistency and pseudo-reference translations in MBR decoding. We include additional sampling strategies such as temperature, top-$k$, nucleus, and epsilon sampling which are orthogonal to ancestral and arithmetic sampling. 

\subsection{Results}
\paragraph{Subsampling} We perform subsampling for all the models and datasets to estimate performance and variance for sampled sequences less than a given $N$ which is 40 for most of our results. For both ancestral and arithmetic sampling, we averaged over multiple runs, sampling sequence sets of lengths $\{ d \in \mathcal{N} \text{ such that } d \, | \, N \}$ from a pool generated by runs with $n$ samples. For arithmetic sampling, we randomly select an offset for each dataset instance and pick indices in fixed intervals to simulate selecting quasi-random samples from a codebook. For ancestral sampling, we randomly sample from all elements in $d$ with replacement. This allowed us to run an experiment for 40 sampled sequences and obtain results for $d \in \{1,2,4,5,8,10,20\}$ sequences. 
\subsubsection{Self-consistency}
We primarily evaluate the performance of the aforementioned sampling techniques with parameters controlled as temperature $=1$, top-$k=40$, and top-$p=0.95$ across $40$ sampled reasoning paths per question-answer pair, using the subsampling results to plot the standard deviation as the shaded region in Figures \ref{fig:gsm8k_analysis} and \ref{fig:csqa_analysis}. We select the majority-voting accuracy for self-consistency as our performance evaluation metric. We also analyze accuracy performance across varying n-gram diversity (controlled by setting temperatures from $\{0.1, 0.3, 0.5, 0.7, 0.9\}$).

\paragraph{Arithmetic reasoning} 
For GSM8K, which consists of open-ended integral answers for grade-school math problems, we performed an 8-shot evaluation as presented in the literature by \citet{kojima2023large}. From Figure \ref{fig:gsm8k_analysis}, we observe significant performance gain from arithmetic sampling over ancestral sampling at 40 sampled sequences: $\mathbf{+4.75\%}$ for Gemma-7B in Figure \ref{fig:gsm8k_analysis} (a) and $\mathbf{+3.25\%}$ for Llama-2-7B in Figure \ref{fig:gsm8k_analysis} (b). The performance v/s diversity analysis from Figure \ref{fig:gsm8k_analysis} (c), shows that as diversity increases, the accuracy increases initially and declines after reaching the maxima at some intermediate diversity level, where arithmetic sampling yields better accuracy than ancestral sampling for all diversity levels.

\paragraph{Commonsense reasoning}
Commonsense QA, a multiple-choice question-answering dataset that necessitates various types of commonsense knowledge to predict correct answers, was utilized for experiments on commonsense reasoning. Similar to arithmetic reasoning, we observe performance gains with arithmetic sampling, as shown in Figure \ref{fig:csqa_analysis}. With Gemma-7B, arithmetic sampling initially trails behind ancestral sampling; however, as the number of sampled sequences surpasses 5, it demonstrates better performance, with a $\mathbf{0.1\%}$ improvement at 40 sampled sequences in Figure \ref{fig:csqa_analysis}(a). In Figure \ref{fig:csqa_analysis} (b), for Llama-2-7B, arithmetic sampling consistently outperforms ancestral sampling across the number of sampled sequences, showing a $\mathbf{1.05\%}$ increase over ancestral at 40 sampled sequences. Conducting performance v/s diversity analysis on Llama-2-7B for 20 sampled reasoning paths with top-$p$ set to 1, as seen in Figure \ref{fig:csqa_analysis} (c), we find that accuracy improves for both sampling methods, with arithmetic reaching a higher global maximum than ancestral and subsequently dominating in performance.

\subsubsection{Machine Translation with MBR}


We ran all the experiments on 1000 machine translation examples with different parameters of temperature $T = \{0.1,1\}$, $\text{top-}p = \{0.8,1\}$, $\text{top-}k = \{30,50\}$, epsilon cutoff $eps= \{0.02,0.05\}$. We selected COMET \cite{rei2020cometneuralframeworkmt, freitag-etal-2022-results} as the evaluation metric. We use subsampling to report the mean COMET score across all the data instances versus the number of sampled sequences = $\{1, 2, 4, 5, 8, 10, 20, 40\}$, plotting the standard deviation as the shaded region in Figure \ref{fig:MBR1}. All the experiments are performed in a zero-shot setting with a simple prompt ``\texttt{Translate the following German/Russian sentence to an English sentence}.``


Figures \ref{fig:MBR1} (a) - (c) show the COMET score of Flan-T5 De-En, MT0 De-En and MT0 Ru-En tasks respectively. Across all COMET score graphs presented (refer Appendix Figure \ref{fig:MBR3} for more plots), it is evident that arithmetic sampling consistently outperforms ancestral sampling as the number of sampled sequences increases. Figures \ref{fig:MBR1} (a), \ref{fig:MBR1} (b), \ref{fig:MBR1} (c) show that arithmetic has \textbf{0.89$\%$}, \textbf{0.48$\%$},  \textbf{0.45$\%$} increase in the mean COMET scores over ancestral at 5 sampled sequences (see Appendix Figure \ref{fig:MBR_table}) without any significant computational overhead \cite{vilnis2023arithmetic}.

We conduct the COMET score vs. n-gram diversity analysis on above tasks by varying temperature $T = \{0.1, 0.3, 0.5, 0.7, 0.9 \}$ and plotting the COMET score for 20 sampled sequences (\ref{fig:MBR2}). As the temperature increases, the diversity of both the sampling techniques increases, thus we chose a higher temperature value of $T=1$ for reporting the results in Figure \ref{fig:MBR1}. The COMET score values for arithmetic outperform those of ancestral, as arithmetic demonstrates greater diversity (\ref{fig:MBR2}). 



We also observe that arithmetic sampling has a lower spread than ancestral (Appendix Figure \ref{fig:MBR_table}). In addition, we conduct the paired t-test at 40 sampled sequences to compare the arithmetic and ancestral COMET scores across 1000 datapoints (Figure \ref{fig:MBR1}). The low p-values indicate that the difference in the mean COMET scores of arithmetic and ancestral sampling is statistically significant.








\subsection{Discussion}

We observe that GSM8K demonstrates greater performance gains compared to Commonsense QA and Machine Translation tasks. This is attributed to nature of the tasks. GSM8K involves open-ended reasoning, where there are no predefined options, and diverse reasoning paths can significantly improve performance by exploring a broader solution space. Arithmetic sampling encourages this diversity, enabling the model to better capture complex reasoning patterns.

In contrast, tasks like Commonsense QA or Machine Translation are more constrained. Commonsense QA has only a limited set of options (e.g., 5 choices), reducing the likelihood of errors and the potential impact of diversity. Similarly, in Machine Translation, translations are typically close to the target sentence, leaving little room for diversity to improve outcomes significantly. This explains why arithmetic sampling shows limited gains in these tasks compared to GSM8K.

\section{Conclusions}
This work investigated the application of arithmetic sampling, a diverse parallelizable sampling technique for multi-sample inference from large language models (LLMs), in chain-of-thought reasoning with self-consistency and machine translation with minimum Bayes risk (MBR) decoding.
Experiments on arithmetic reasoning and commonsense reasoning demonstrated that arithmetic sampling improves performance as the number of sampled sequences increases, yielding $3$-$5\%$ and $1.05\%$ point increases in accuracy, respectively, over ancestral sampling at $40$ sampled sequences.
In machine translation, arithmetic sampling consistently outperformed ancestral sampling, with improvements ranging from $0.45\%$ to $0.89\%$ in mean COMET scores for $5$ sampled sequences across different language pairs and parameter settings. Arithmetic sampling also exhibited a lower standard deviation compared to ancestral sampling.
The superior performance of arithmetic sampling with multi-sample inference based decoding methods like self-consistency and MBR decoding is due to its ability to generate more diverse samples, enhancing LLM performance in tasks like chain-of-thought reasoning and machine translation without any computational overhead.

\section{Limitations}
\begin{enumerate}
    \item \textbf{Limited to random vocabulary ordering:} As the unit interval associated with arithmetic sampling corresponds to an ordering of the vocabulary tokens (from which sequences are decoded via the code points), we limited ourselves to randomizing the vocabulary ordering and then performing arithmetic sampling. At the expense of some constant overhead, creating an ordering of the vocabulary that follows some measure of semantic similarity may show greater diversity (and potentially, better performance) as the uniform lattice of code points would now be corresponding to semantically dissimilar vocabulary tokens more often -- rendering this idea an interesting future direction. 
    \item \textbf{Limited to one dimension:} In arithmetic sampling, tokens are limited in one dimension space with softmax sampling. It may be beneficial to consider higher dimensional embeddings of the tokens with box embeddings \cite{vilnis2018probabilistic}. Diverse sampling in high dimensional embedding spaces may involve advanced sampling techniques like Quasi-Monte Carlo \cite{caflisch1998monte}.
    \item \textbf{Difference between sampling and maximizing:} Arithmetic sampling was formulated to generate parallel diverse sequences, which differs from strategies that find the most probable answer like beam search. Our application of arithmetic sampling is limited to multi-sample decoding strategies that require diversity. It may be an interesting future work to extend arithmetic sampling for MAP decoding.
\end{enumerate}

\clearpage
\bibliography{iclr2025_conference}
\bibliographystyle{iclr2025_conference}

\newpage
\appendix

\clearpage
\section{Appendix}
\label{sec:appendix}
\subsection{Definition of sample diversity}  \label{app: A1}
\text{The $n$-gram diversity score is defined as } $d = \sum_{n=1}^4 d_n$ \text{ where}

$d_n = \frac{\text{\# of unique $n$-grams in $N$ translations}}{\text{total \# of $n$-grams in $N$ translations}}$.
\subsection{Machine Translation with MBR}

\begin{figure*}[h]
\centering
\resizebox{0.8\linewidth}{!}{%
    \begin{tabular}{cccc}
     \captionsetup[subfloat]{font=Huge,labelfont=Huge}
         \subfloat[De-En task: Flan-T5 ]{{\includegraphics[width=1\linewidth]{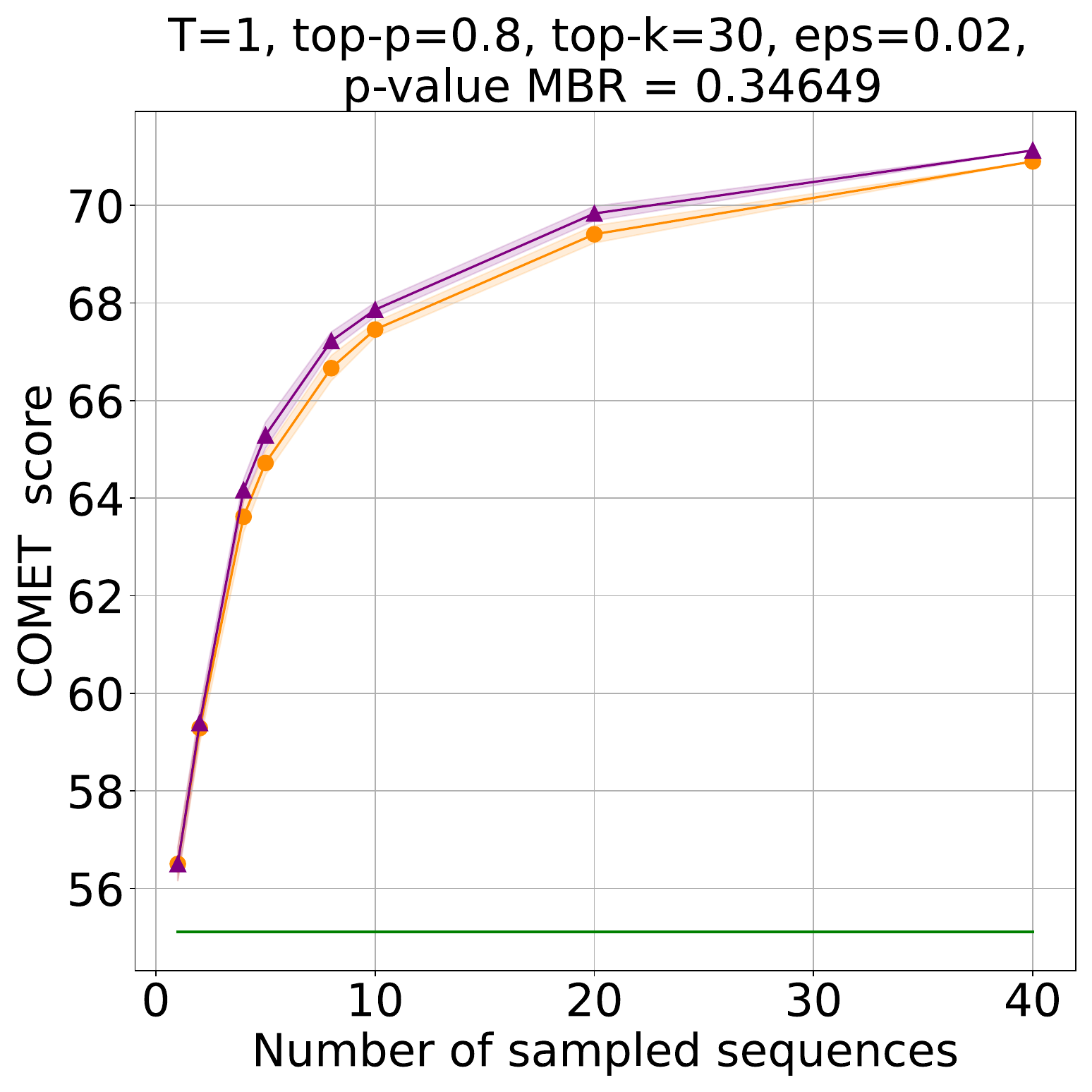} }} &
          \captionsetup[subfloat]{font=Huge,labelfont=Huge}
        \subfloat[De-En task: MT0 ]{{\includegraphics[width=1\linewidth]{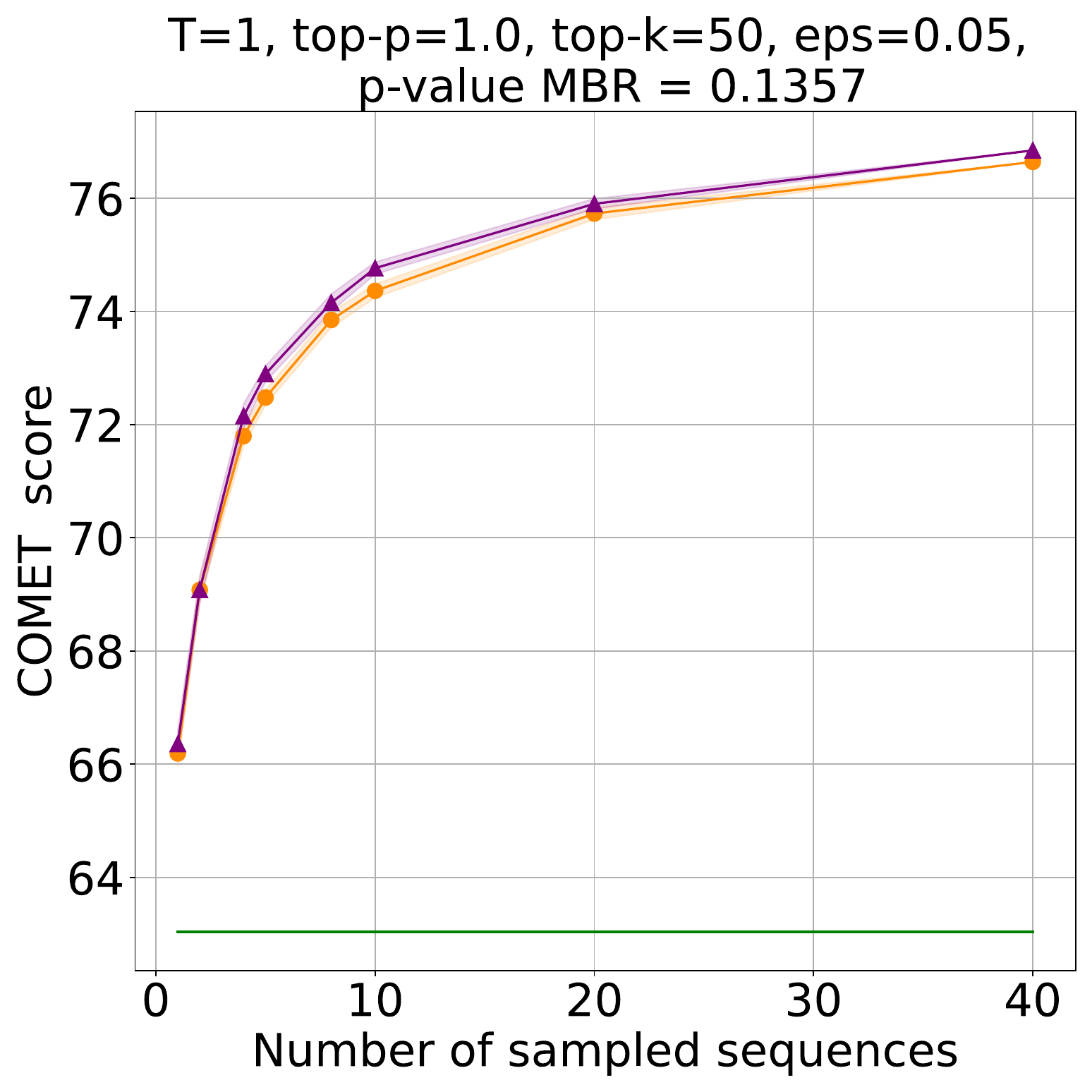} }} &
         \captionsetup[subfloat]
         {font=Huge,labelfont=Huge} 
         \subfloat[Ru-En task: MT0 ]{{\includegraphics[width=1\linewidth]{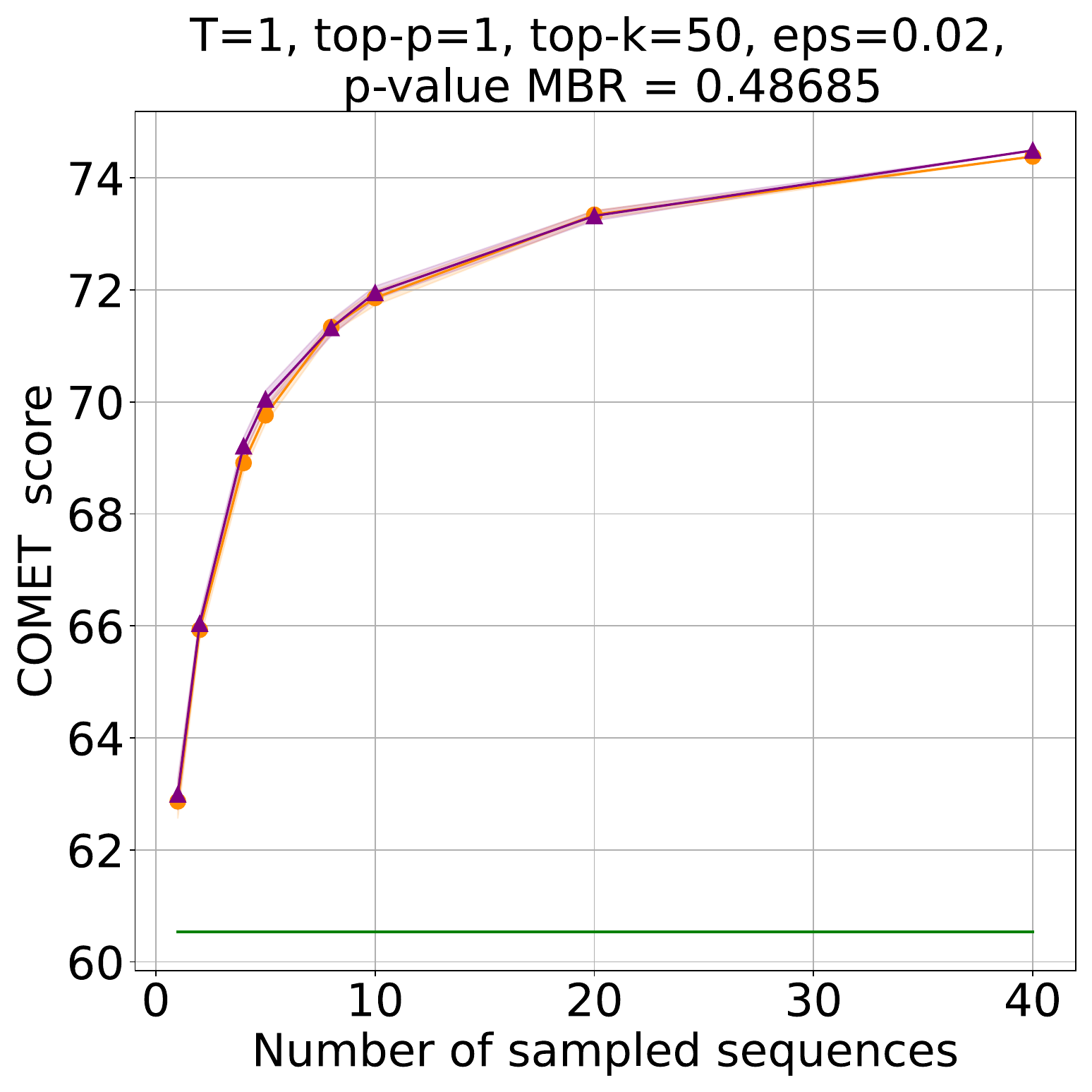} }} &
         \captionsetup[subfloat]
         {font=Huge,labelfont=Huge} 
    \end{tabular}
}
      \begin{tikzpicture}
        \begin{customlegend}[
            legend columns=3,
            legend style={
                align=left,
                column sep=2ex
            },
            legend entries={ Ancestral sampling (MBR), Arithmetic sampling (MBR), Greedy}
        ]
            \addlegendimage{mark=*,solid,color=acolor}
            \addlegendimage{mark=triangle*,mark size=3pt,solid,color=bcolor}
            \addlegendimage{mark=circle*,solid,color=ccolor}
        \end{customlegend}
    \end{tikzpicture}   
    \caption{COMET vs. \#sampled sequences on Flan-T5, MT0}
    \label{fig:MBR3}
    \vspace{-10pt}
\end{figure*}

In this section, we present more plots on De-En, Ru-En task for Flan-t5 and MT0 models across different parameter setting as shown in the Figure \ref{fig:MBR3}.

In addition, we also report the mean and standard deviation of performance (COMET) for arithmetic and ancestral sampling for the best performing parameter setting in Figure \ref{fig:MBR_table}.

\begin{figure*}[!h]
\centering
\resizebox{\linewidth}{!}{%
\begin{tabular}{cccc}
    \captionsetup[subfloat]{font=Large,labelfont=Large}
    \subfloat[De-En task: Flan-T5 (T = 1, top-p = 1.0, top-k = 50, eps = 0.02 )]
    {{\begin{tabular}{ccc}
        \toprule
        \textbf{\#Sampled Seq} & \textbf{Arithmetic} & \textbf{Ancestral} \\
        \midrule

1  & 55.19 $\pm$ 0.35 & 55.39 $\pm$ 0.36 \\ 
2 & 58.38 $\pm$ 0.21 & 58.51 $\pm$ 0.29 \\ 
4 & 63.24 $\pm$ 0.19 & 62.72 $\pm$ 0.33 \\ 
5 & 64.48 $\pm$ 0.18 & 63.91 $\pm$ 0.21 \\ 
8 & 66.27 $\pm$ 0.14 & 65.82 $\pm$ 0.21 \\ 
10  & 67.19 $\pm$ 0.12 & 66.64 $\pm$ 0.24 \\ 
20 & 69.11 $\pm$ 0.12 & 68.80 $\pm$ 0.18 \\ 
40 & 70.46 $\pm$ 0.00 & 70.13 $\pm$ 0.00 \\ 

        \bottomrule
    \end{tabular} }}
    &
    \captionsetup[subfloat]{font=Large,labelfont=Large}
    \subfloat[De-En task: MT0 (T = 1, top-p = 0.8, top-k = 30, eps = 0.02 )]
    {{\begin{tabular}{ccc}
        \toprule
        \textbf{\#Sampled Seq} & \textbf{Arithmetic} & \textbf{Ancestral} \\
        \midrule

        1  & 65.74 $\pm$ 0.28 & 65.71 $\pm$ 0.18 \\ 
2  & 68.75 $\pm$ 0.25 & 68.58 $\pm$ 0.22 \\
4  & 71.85 $\pm$ 0.22 & 71.59 $\pm$ 0.19 \\ 
5  & 72.72 $\pm$ 0.13 & 72.37 $\pm$ 0.15 \\ 
8  & 74.00 $\pm$ 0.13 & 73.75 $\pm$ 0.13 \\
10 & 74.62 $\pm$ 0.11 & 74.27 $\pm$ 0.16 \\
20 & 75.90 $\pm$ 0.07 & 75.57 $\pm$ 0.12 \\ 
40 & 76.82 $\pm$ 0.00 & 76.50 $\pm$ 0.00 \\ 

        \bottomrule
    \end{tabular} }}
    &
    \captionsetup[subfloat]{font=Large,labelfont=Large}
    \subfloat[Ru-En task: MT0 (T = 1, top-p = 1.0, top-k = 50, eps = 0.05 )]
    {{\begin{tabular}{ccc}
        \toprule
        \textbf{\#Sampled Seq} & \textbf{Arithmetic} & \textbf{Ancestral} \\
        \midrule

    1  & 64.80 $\pm$ 0.23  & 64.48 $\pm$ 0.23 \\ 
    2  & 67.55 $\pm$ 0.19  & 67.41 $\pm$ 0.27 \\ 
    4  & 70.29 $\pm$ 0.14  & 70.01 $\pm$ 0.13 \\ 
    5  & 71.06 $\pm$ 0.12  & 70.74 $\pm$ 0.18 \\ 
    8  & 72.32 $\pm$ 0.10  & 71.99 $\pm$ 0.14 \\ 
    10 & 72.72 $\pm$ 0.11  & 72.51 $\pm$ 0.12 \\ 
    20 & 73.85 $\pm$ 0.09  & 73.74 $\pm$ 0.08 \\ 
    40 & 74.82 $\pm$ 0.00  & 74.71 $\pm$ 0.00 \\ 
        \bottomrule
    \end{tabular} }}
    &
\end{tabular}
}
\captionsetup{singlelinecheck=false}
\caption{Tables showing mean$\pm$std of mean COMET for 20 runs using arithmetic and ancestral sampling,  based on the number of sampled sequences}
\label{fig:MBR_table}
\end{figure*}

\end{document}